\pdfoutput=1
\documentclass[11pt]{article}
\usepackage[]{acl}
\usepackage{times}
\usepackage{latexsym}
\usepackage{booktabs}

\usepackage[T1]{fontenc}

\usepackage{graphicx}
\usepackage{epstopdf}

\usepackage[utf8]{inputenc}

\usepackage{microtype}

\title{HIT at SemEval-2022 Task 2: Pre-trained Language Model for Idioms Detection}

\author{
Zheng Chu$^\dag$$^\ddag$,
Ziqing Yang$^\ddag$,
Yiming Cui$^\dag$$^\ddag$,
Zhigang Chen$^\ddag$,
Ming Liu$^\dag$ \\
{$^\dag$Research Center for SCIR, Harbin Institute of Technology, Harbin, China} \\
{$^\ddag$State Key Laboratory of Cognitive Intelligence, iFLYTEK Research, China} \\
$^\dag$\tt \{zchu,ymcui,mliu\}@ir.hit.edu.cn \\
$^\ddag$\tt\{zhengchu,zqyang5,ymcui,zgchen\}@iflytek.com }

\begin{document}

\maketitle
\begin{abstract}
The same multi-word expressions may have different meanings in different sentences. They can be mainly divided into two categories, which are literal meaning and idiomatic meaning.
Non-contextual-based methods perform poorly on this problem, and we need contextual embedding to understand the idiomatic meaning of multi-word expressions correctly.
We use a pre-trained language model, which can provide a context-aware sentence embedding, to detect whether multi-word expression in the sentence is idiomatic usage.

\end{abstract}

\section{Introduction}
The goal of the SemEval-2022 Task2 \cite{semeval} SubtaskA is to detect whether a multi-word expression in a sentence is idiomatic in usage. It is a multilingual task and consists of three languages: English, Portuguese and Galician.

\par Multi-word expressions (MWEs) are expressions that consist of at least two words and are syntactically or semantically specific. The semantics of MWEs are usually divided into two types, (i) the combination of literal meanings of each word in the phrase or (ii) inherent usages (e.g., idiomatic meaning). Understanding the semantic meaning of a sentence requires the correct identification of the MWE in the sentence. Table \ref{table_date_example} contains one case for each of the two usages \cite{astitchinlm}.

Traditional non-contextual word embedding models, such as word2vec \cite{DBLP:journals/corr/abs-1301-3781}, perform poorly at this task. Simple superposition of non-contextually word embeddings does not correctly express the semantics of idiomatic phrases. Therefore, contextual embedding models \cite{conneau-etal-2020-unsupervised,DBLP:conf/naacl/DevlinCLT19} are required to correctly understand the meaning of multi-word expressions in idiomatic usage.\par

We used large-scale cross-lingual pre-trained language models, multilingual BERT \cite{DBLP:conf/naacl/DevlinCLT19} and XLM-RoBERTa \cite{conneau-etal-2020-unsupervised}, with a softmax classifier on top of the pre-trained LM to train a classification model.  The training data are processed before training, and regularization dropout \cite{DBLP:journals/corr/abs-2106-14448}, adversarial training \cite{DBLP:conf/iclr/MiyatoDG17, DBLP:conf/iclr/MadryMSTV18} are used in the training process. In addition, we observed the training data and found an interesting phenomenon that we can get better results by post-processing after the training using heuristic rule.

\section{Background}

\subsection{Task Description}
Task 2 contains two subtasks, SubtaskA is idiom detection, and SubtaskB is similarity scoring of texts containing idioms. This article focus on SubtaskA. SubtaskA contains two settings, zero-shot and one-shot.
\begin{itemize}
    \item Zero-shot: Multi-word expressions that appear in test data do not appear in training data.
    \item One-shot: Every multi-word expressions that appeared in the test data appeared in training data at least once.
    \item Data Restriction: Under zero-shot setting, we can only use zero-shot training data, and we can use both zero-shot and one-shot training data under one-shot setting. Test data is same for both settings.
\end{itemize}
\begin{table}[]
\small \centering
\begin{tabular}{@{}ll@{}}
\toprule
\textbf{Literal} &
  \begin{tabular}[c]{@{}l@{}}When removing a \underline{\textbf{big fish}} from a net,\\ it should be held in a manner that\\ suports the girth.\end{tabular} \\ \midrule
\textbf{Idiomatic} &
  \begin{tabular}[c]{@{}l@{}}It was still a respectable finish for\\ both Fadol and Nayre, who were ranked\\ outside the top 500 in the world but\\ caught some \underline{\textbf{big fish}} along the way\end{tabular} \\ \bottomrule
\end{tabular}
\caption{Examples of idiomatic and non-idiomatic usage}
\label{table_date_example}
\end{table}

\subsection{Data Details}
In this section, we will describe the characteristics of the training data and test data. The official data includes eight columns, which are DataID, Language, MWE, Setting, Previous, Target, Next, Label. 
\begin{figure*}[!h]
\centering
\includegraphics[scale=0.45]{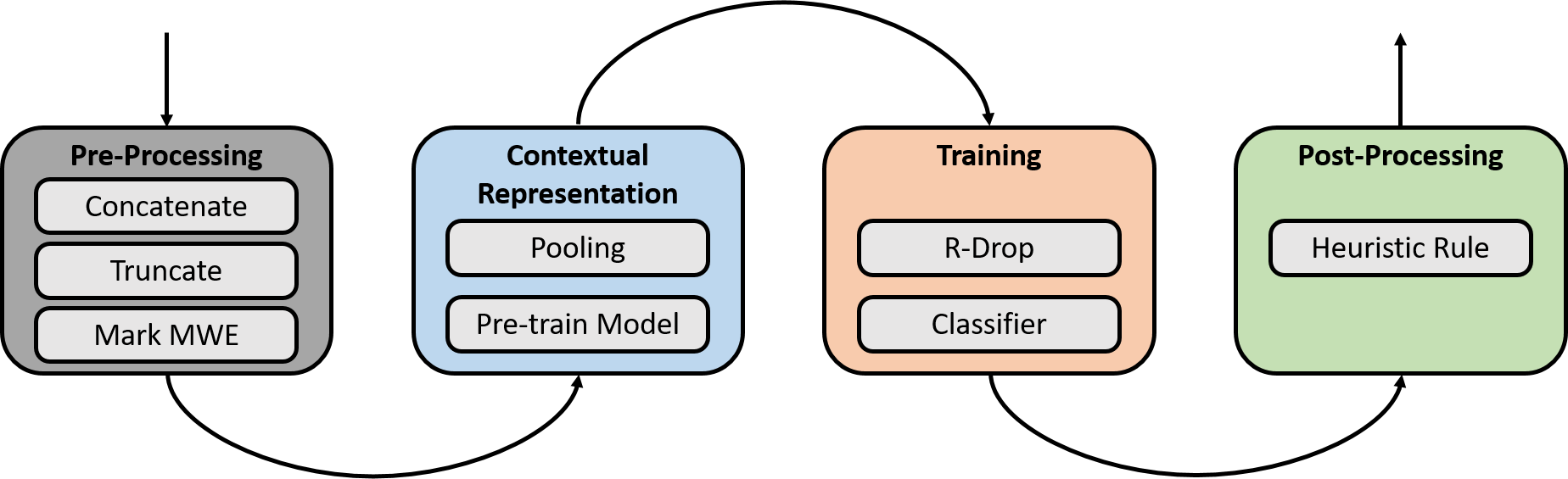}
\caption{System flow diagram}
\label{figure_system_flow}
\end{figure*}

\subsubsection{Language}
The test data includes 916, 713, 713 entries in English, Portuguese and Galician, respectively.  
\par There are only English and Portuguese examples in the training data of zero-shot setting, which means that in the testing phase, the model requires zero-shot transfer of Galician with the learned knowledge on other two languages. Any MWE in zero-shot training data will not appear in the test data.
\par In one-shot training data, there are 73 entries in each of the three languages, which is relatively small compared to zero-shot training data. Any MWE in the one-shot training data will appear in the test data.

\subsubsection{Data Length}
We counted the average length of the data to facilitate appropriate truncation when using a pre-trained model.
\par The average, median, max length of target sentences after tokenizer corresponding to the pre-train model are 42.6, 195, 40, respectively. Over 90\% of sentences are 64 or less in length.
\par The average, median, max position that MWE occurs in sentences after tokenizer is 18.9, 89, 15, respectively, and over 90\% of sentences are 37 or less in position.

\subsection{Related Work}
So far, there has been extensive research about idioms detection. \citet{10.1162/tacl_a_00442} propose a multi-stage neural architecture with attention flow. 
\citet{garcia-etal-2021-assessing, garcia-etal-2021-probing} probs idiomaticity in vector space and propose NCTTI dataset. \citet{do-dinh-etal-2018-killing} propose a multi-task learning method. \citet{astitchinlm} present a multilingual idiom detection dataset, which will be used as this SemEval-2022 idioms detection track.
\par Pre-trained word embedding can capture syntactic and semantic information from large amounts of unlabeled data, which has been a standard part of natural language processing task \cite{DBLP:journals/corr/abs-1301-3781, pennington-etal-2014-glove}. However, each of these methods can only obtain a fixed, non-contextual vector representation for each word which makes it difficult to convey the correct meaning of polysemous words. Due to the disadvantages of non-contextual embedding, recent work has begun to 
focus on contextual embedding, typical cases are context2vec \cite{melamud-etal-2016-context2vec}, ELMo \cite{peters-etal-2018-deep}, BERT \cite{DBLP:conf/naacl/DevlinCLT19}.
\par The most commonly used in contextual embedding is the pre-trained language model \cite{DBLP:conf/naacl/DevlinCLT19, conneau-etal-2020-unsupervised}. These models perform self-supervised training through mask language modeling, next sentence predicting, and other objectives in hundreds of millions of unlabeled datas. Benefiting from multilingual training data, these models have cross-language capabilities. Pre-trained models are gradually taking the place of pre-trained word embeddings as the new paradigm for natural language processing.

\section{System Overview}

Figure \ref{figure_system_flow} depicts the flow chart of the whole system. We first preprocess the data, tokenizing and then feed into a pre-trained model to get hidden states. Apply some pooling method to get fixed length sentence representation to train a softmax binary classifier. After that, the prediction results of the model are post-processed. In the training process, contrastive learning, adversarial training, regularized dropout, etc. are used. Table \ref{tabel_train_example} is then used as an example to introduce the data preprocessing process of the system. 

\subsection{Baseline}
The baseline method below refers to the method in paper \cite{astitchinlm}.
\begin{itemize}
    \item In the zero-shot setting, Multilingual BERT is trained on zero-shot data, using the context without idiom as an additional feature.
    \item In the one-shot setting, Multilingual BERT is trained on combination of the zero-shot and one-shot data, excluding the context and adding the idiom as an additional feature.
\end{itemize}

\subsection{Data Preprocessing}
\subsubsection{Truncation}
The data provides the context of the target sentence. We only use target sentences for training because we found that if we concatenate previous, target, and next together, the sentence length will be too long, which slows down training and harms performance. We guess that to distinguish whether the MWE is an idiomatic usage, we only need to focus on words near the MWE, too long sentences will introduce unnecessary distractions. 
\par According to the length statistics in the previous chapter, 128 is used as the maximum truncation length of the pre-trained model, which can ensure that most sentences will not be truncated and keep the sentence length as small as possible.
\subsubsection{MWE Marking}
Following the baseline method, we use the tokenizer's [SEP] token to mark the MWE in the sentence. Unlike the baseline method, we only mark MWEs without deformation. Proper nouns are usually non-idiomatic usage, and are often deformed. Pre-trained models can recognize proper nouns well, so we do not mark the deformed MWEs. The results also show that this gives better performance. 
\par Example in Table \ref{tabel_train_example}, MWE ‘milk tooth’ in sentence "Her latest pamphlet \underline{\textbf{Milk Tooth}}, published by Rough Trade Books, is a collection of thwarted escape plans for a too-heavy world" is capitalized, according to our rules, the MWE in this sentence is not marked. If the MWE in the sentence is not deformed, the [SEP] token will be used to mark the MWE, just like [SEP]\underline{\textbf{milk tooth}}[SEP].

\subsection{Model}
\subsubsection{Pre-trained Model}
We tried different multilingual pre-trained models, including mBERT and XLM-RoBERTa, and XLM-RoBERTa consistently outperforms mBERT. In addition, we also try to use different size models, including mBERT-base-cased, XLM-RoBERTa-base, XLM-RoBERTa-large, and bigger models lead to better performance.
\subsubsection{Classifier}
Different pooling methods are used for hidden states of different layers, including mean pooling, max pooling, [CLS], and token-level pooling. The results show that different pooling methods have little effect on the results. For simplicity, [CLS] is used as the sentence representation. 
\par For token-level pooling, we pool the vectors of MWE positions in hidden states to obtain the final sentence representation, which harms the performance.
\par After pooling on the pre-trained model, fixed-length sentence representation is obtained. This is followed by a full connection layer with dropout \cite{DBLP:journals/jmlr/SrivastavaHKSS14} and a softmax classifier.

\subsubsection{Regularized Dropout}
\begin{figure}[!h]
\centering
\includegraphics[scale=0.5]{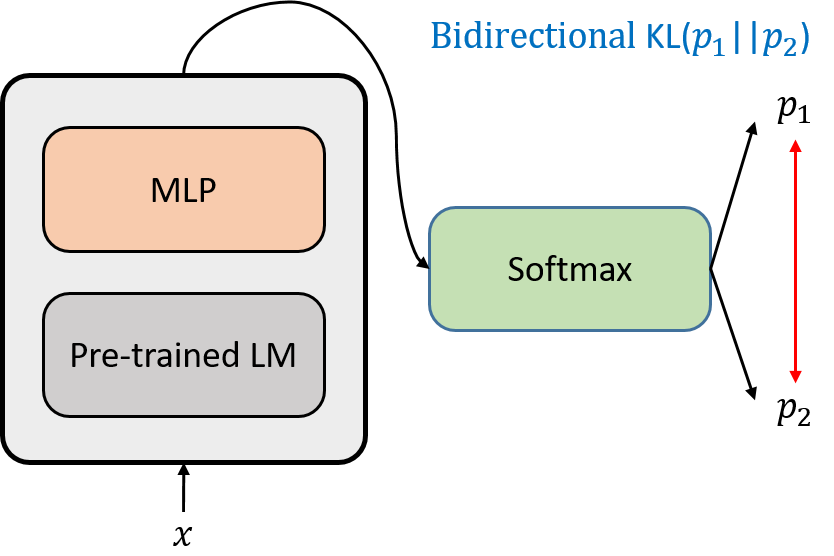}
\caption{Regularized Dropout}
\label{figure_rdrop}
\end{figure}

\par Deep neural networks usually use dropout \cite{DBLP:journals/jmlr/SrivastavaHKSS14}, but the use of dropout introduces inconsistency between train and inference. R-drop \cite{DBLP:journals/corr/abs-2106-14448} is a means of regularization in the training process to mitigate this inconsistency.
\par A sample output through the pre-trained model, MLP and softmax can be regarded as a probability distribution. R-drop performs two independent forward calculations for each sample, obtaining two outputs probability distribution. Due to the dropout, these two outputs will be slightly different, introducing inconsistency. To mitigate this, the bi-directional KL-divergence is calculated as a penalty between these two outputs probability distributions. In the following equation, $y_i$ represents the label, $x_i$ represents the 
input, and the superscript represents two independent forward operations. $p_i^1, p_i^2$ represent probability distribution obtained from two independent forwards.
 
\begin{equation}
\mathcal{L}_{CE}^{i} = CE(x_i^1, y_i) \ + \ CE(x_i^2, y_i)
\end{equation}
\begin{equation}
\mathcal{L}_{KL}^{i} = \frac{1}{2}(\mathcal{D}_{KL}(p_i^1||p_i^2) +\mathcal{D}_{KL}(p_i^2||p_i^1))
\end{equation}
\begin{equation}
\mathcal{L}^{i} = \mathcal{L}^{i}_{CE} + \alpha \cdot \mathcal{L}^{i}_{KL}
\end{equation}

\subsubsection{Post Processing}
\label{sec:postprocess}
We found an interesting phenomenon in the training data. Some MWEs in one-shot training data have only one category label, and most of these MWEs corresponding to entries in the dev data have the same label as in the training data. Some proper nouns are labeled with idiomatic meanings, however some with literal meanings. These labeling inconsistencies may cause problems in the learning of the model, so we design a heuristic rule. On top of the model prediction results, if there is only one label for a certain MWE in the training set, then replace all the predictions for that MWE in the test set with whichever label appears in the training set.

\begin{table}[]
\centering
\small
\begin{tabular}{@{}ll@{}}

\toprule
\textbf{MWE}      & milk tooth                                                                                               \\ \midrule
\textbf{Previous} & \begin{tabular}[c]{@{}l@{}}A ritual sacrifice from the 19th \\ century is vividly relieved.\end{tabular} \\ \midrule
\textbf{Target} &
  \multicolumn{1}{l}{\begin{tabular}[c]{@{}l@{}}Her latest pamphlet \textbf{\underline{Milk Tooth}}, \\ published by Rough Trade Books, \\ is a collection of thwarted \\ escape plans for a too-heavy world.\end{tabular}} \\ \midrule
\textbf{Next} &
  \multicolumn{1}{l}{\begin{tabular}[c]{@{}l@{}}In these poems of trauma and \\ transformation, the present throbs\\  with unfinished histories.\end{tabular}} \\ \midrule
\textbf{Label}    & 1                                                                                                        \\ \bottomrule
\end{tabular}
\caption{Training data example (useless columns have been removed).}
\label{tabel_train_example}
\end{table}

\section{Experiments} 
\subsection{Hyperparameters}
We use the Huggingface Transformers \cite{DBLP:journals/corr/abs-1910-03771} implementation of mBERT and XLM-RoBERTa. During the training, the learning rate schedule strategy is warmup of first 10\% steps with cosine learning rate decay in rest steps. For zero-shot and one-shot, we use learning rates of 1e-5 and 3e-5, respectively, and one-shot is continued training on the well-performing zero-shot model. The mini-batch size is 32. The coefficient of R-drop is chosen from 0, 1, 2, 4. We train a total of 20 epochs and save the best-performing checkpoints on the development set. All models are trained on one single NVIDIA Tesla V100 GPU.
\subsection{Evaluation Metrics}
SubtaskA is evaluated using the Macro F1 score between the gold labels and model predictions.
\begin{equation}
Macro \ F1 = 2 \cdot \frac{precision \times recall}{precision + recall}
\end{equation}
\subsection{Hyperparameters Selection}
We compare different pre-trained models, max length, and whether to use MWE and contexts.
\subsubsection{Context and Idiom}
\begin{table}[]
\centering
\begin{tabular}{@{}lcc@{}}
\toprule
Model                      & Zero-shot & One-shot \\ \midrule
mBERT$_{base}$ w/ C w/ I   &     74.90    &    84.78     \\
mBERT$_{base}$ w/ C w/o I  &     70.59    &    76.98      \\
mBERT$_{base}$ w/o C w/ I  &     \textbf{75.31}    &    \textbf{85.76}     \\
mBERT$_{base}$ w/o C w/o I &     70.76    &    82.59   \\ \bottomrule
\end{tabular}
\caption{Context and idiom effects on the development set results. C: context. I: Idiom (Macro F1 $\times$ 100).}
\label{table_context_idiom}
\end{table}
We compare the performance of the model with and without context in Table \ref{table_context_idiom}, the effect of marking idioms on the results. Moreover we conduct experiments on BERT$_{base}$, using [CLS] as pooling, with the max sentence length set to 128. The method of marking MWEs here follows the baseline.
\par Experiments show that ignoring context and mark idioms gives better results, and this setting will be continued for future experiments.

\begin{table}[]
\centering
\begin{tabular}{@{}lccc@{}}

\toprule
Model                   & Max Length &Zero-shot & One-shot \\ \midrule
mBERT$_{base}$    &  128   & \textbf{76.31}       &   \textbf{87.97}       \\
mBERT$_{base}$    &  192   & 75.62       &   87.96       \\
mBERT$_{base}$    &  256   & 75.64       &   86.24       \\ \bottomrule
\end{tabular}
\caption{Effect of different maximum sentence lengths on the development set results (Macro F1 $\times$ 100).}
\label{table_length}
\end{table}

\subsubsection{Max Sentence Length}

In this section, a comparison is made between the cases with different maximum sentence lengths. The model follows the previous setting, with idioms marked and context ignored, using first-last-avg as pooling for training. 
\par Our original hypothesis is that performance and speed are a trade-off as the maximum sentence length increases. In contrast, Table \ref{table_length} shows that a maximum sentence length of 128 is sufficient in terms of speed and performance. 
We do not test a smaller maximum sentence length because further reduction might cause parts of the sentence to be truncated, harming the performance.

\subsubsection{Pre-trained Model}

\begin{table}[]
\centering
\begin{tabular}{@{}lcc@{}}
\toprule
Model           & Zero-shot & One-shot \\ \midrule
mBERT$_{base}$  & 75.31       &    87.97                \\
XLM-R$_{base}$   & 76.99        &     89.15               \\
XLM-R$_{large}$  &  \textbf{78.17}       &     \textbf{91.84}                \\ \bottomrule
\end{tabular}
\caption{Effect of different pre-trained models on the development set (Macro F1 $\times$ 100).}
\label{table_pretrain}
\end{table}
We compared mBERT$_{base}$, XLM-RoBERTa$_{base}$, and XLM-RoBERTa$_{large}$, and results in Table \ref{table_pretrain} demonstrated that XLM-RoBERTa outperformed mBERT, and the large model performed better than base model. In addition, our final submission results were obtained using XLM-R$_{large}$.
\par From the results, we found that the performance improvement is evident as the model size grows, and there is no bottleneck yet. Therefore, increasing the model size may be a simple and effective way.

\section{Results}
\subsection{System Performance}
Our final results use model fusion on thirteen models. The zero-shot model finished fourth with an F1 of 77.15, and the one-shot model finished first with an F1 of 93.85.
\subsection{Ablation Study}
Table \ref{table_ablation} provides the results of the ablation study. The baseline of the ablation experiment follows the hyperparameters of the experiment chapter, except that the model is replaced with XLM-R$_{large}$. 
\subsubsection{Mark MWE}
\par First, we change the method of marking MWE in the baseline. We use [SEP] token for tagging only if the MWE in the sentence is the same as the MWE provided in the data. The reason for this is that the organizer's rule is to label proper nouns as literal meaning, and proper nouns are usually deformed with initial capitalization. The pre-trained model can distinguish proper nouns well under the training of a large amount of corpus.
\subsubsection{Adversarial Training}
Adversarial training \cite{DBLP:conf/iclr/MiyatoDG17, DBLP:conf/iclr/MadryMSTV18} is a way to enhance the robustness of neural networks by adding small perturbations to the samples to interfere with the predictions of the model. In our experiments, the results in Table \ref{table_ablation} show that adversarial training significantly improves performance.
\subsubsection{R-drop}
R-drop is a regularization tool that aims to maintain the consistency of model prediction and training while using dropout by adding bi-directional KL-divergence as a penalty term. After adding R-drop, the performance is significantly improved, exceeding the adversarial training. In Table \ref{table_ablation}, under zero-shot setting, the relative improvement is 2.17 and 0.56 compared to baseline and adversarial training, respectively, and this improvement is 1.07 and 0.44 under one-shot setting, respectively.
\subsubsection{Heuristic Rule}
We used the heuristic rule mentioned in Section \ref{sec:postprocess} for replacement under the one-shot setting, and as shown in the Table \ref{table_ablation}, the relative improvement is 0.82 percentage points.

\begin{table}[]
\centering
\begin{tabular}{@{}lcc@{}}
\toprule
Model      & Zero-shot & One-shot \\ \midrule 
XLM-R$_{large}$ &  78.05      &    91.02    \\
+mark MWE  &   78.17        &    91.84      \\
\ \ +contrastive pre-train         &    -       &     89.95     \\
\ \ +contrastive auxiliary        &    76.30       &     88.05     \\
\ \ +AEDA       & 79.09     &  89.76\\
\ \ +AT        &    79.78       &     92.47     \\
\ \ +R-drop    &   \textbf{80.34}        &     \textbf{92.91}     \\
\ \ \ \ +post-processing  & - & \textbf{93.73}  \\

 \bottomrule
\end{tabular}
\caption{Ablation experiments on the development set (Macro F1 $\times$ 100).}
\label{table_ablation}
\end{table}
\subsubsection{Negative Results}
\textbf{Contrastive Learning}. Recently, contrastive learning has been a  hot topic in NLP, especially in sentence representation learning. Inspired by SimCSE \cite{gao-etal-2021-simcse}, we use contrastive learning before and during training classification, using the data from subtaskA for contrastive pre-train and apply contrastive loss during training as an auxiliary training objectives, respectively. Unfortunately, as shown in Table \ref{table_ablation}, both of these methods make the results much worse.

\noindent \textbf{Data Augmentation}. Due to lack of training data, we use data augmentation for expansion. We used AEDA \cite{AEDA} as a means of data augmentation, which is a straightforward data augmentation, by adding punctuation marks to the sentences. The results showed that a particular improvement was achieved under zero-shot setting, but a decrease was achieved under one-shot setting.

\section{Conclusion}
We continuously improve the model performance by improving the MWE marking method, using larger pre-trained models, adding regularization terms and heuristic rules. Inspired by pre-trained models, we can find that future work needs to focus more on external data besides training data, especially data with specific idioms, such as idiom dictionaries, because the size of the external data is much larger than the training data, which has not been fully exploited.

\bibliography{custom}

\appendix

\end{document}